\documentclass[letterpaper, 10 pt, conference]{ieeeconf} 

\usepackage{hyperref}
\usepackage{url}
\usepackage{graphics} 
\usepackage{times} 
\usepackage{booktabs}       
\usepackage{amsfonts}       
\usepackage{nicefrac}       
\usepackage{microtype}      
\usepackage{algorithm,multirow,xcolor}
\usepackage{algorithmicx}
\usepackage{algpseudocode}
\usepackage{mathtools}
\usepackage{graphicx}
\usepackage{cases}
\usepackage{array}
\usepackage{color}
\usepackage{float}
\usepackage{amssymb}

\usepackage{wrapfig}
\usepackage{subfigure}
\usepackage{amsmath}
\usepackage{soul}

\usepackage{graphicx} 
\usepackage{subfigure}
\usepackage{epstopdf}

\newcommand{\define}{\triangleq}

\newtheorem{theorem}{Theorem}
\newtheorem{definition}{Definition}

\newtheorem{assumption}{Assumption}

\IEEEoverridecommandlockouts 
\title{\LARGE \bf
Reinforcement Learning for Orientation Estimation \\Using Inertial Sensors with Performance Guarantee
}

\author{Liang Hu$^{1*}$,  Yujie Tang$^{2*}$, Zhipeng Zhou$^{2}$ and Wei Pan$^{2}$
\thanks{*Equal Contribution. }
\thanks{$^{1}$L.~Hu is with the School of Computer Science and Electronic Engineering, University of Essex, UK.}%
\thanks{$^{2}$Y.~Tang, Z.~Zhou and W.~Pan are with  the Department of Cognitive Robotics, Delft University of Technology, Netherlands. For Correspondence: \texttt{\small wei.pan@tudelft.nl}.}
}

\begin{document}

\maketitle
\thispagestyle{empty}
\pagestyle{empty}

\begin{abstract}
 This paper presents a deep reinforcement learning (DRL) algorithm for orientation estimation using inertial sensors combined with a magnetometer. Lyapunov's method in control theory is employed to prove the convergence of orientation estimation errors. The estimator gains and a Lyapunov function are parametrised by deep neural networks and learned from samples based on the theoretical results. The DRL estimator is compared with three well-known orientation estimation methods on both numerical simulations and real dataset collected from commercially available sensors. The results show that the proposed algorithm is superior for arbitrary estimation initialisation and can adapt to a drastic angular velocity profile for which other algorithms can be hardly applicable. To the best of our knowledge, this is the first DRL-based orientation estimation method with an estimation error boundedness guarantee.
\end{abstract}

\section{Introduction}
Orientation estimation is essential in robotics, navigation, control, and human motion analysis \cite{barfoot2017state, zhou2008human, kok2017using}. Recently, orientation estimation has been dramatically advanced by the development of accurate sensors. Multiple sensors are usually combined to estimate the orientation, i.e., sensor fusion. Depending on the availability of sensors and applications, various sensor fusion techniques have been proposed, e.g., the inertial measurement units (IMU) and magnetometer \cite{sabatini2006quaternion, marins2001extended, valenti2015keeping}, the magnetometer and camera \cite{wang2016keyframe}, and the IMU and visual sensor \cite{clark2017vinet, qin2018vins}, etc. In this paper, we focus on orientation estimate using inertial sensors and a magnetometer.

The estimation algorithms can be summarised into three categories: (1) Bayesian estimation, (2) optimisation and (3) deep learning. In Bayesian estimation, the well-known extended Kalman filter (EKF) and the unscented Kalman filter (UKF) were used to estimating the orientation \cite{sabatini2006quaternion, marins2001extended,kraft2003quaternion}. The key idea is to approximate the orientation states by a Gaussian distribution based on the linearisation technique and the deterministic sampling technique. Furthermore, the complementary filter was developed based on the EKF, which exploits the complementary characteristics of gyroscopes and that of accelerometer and magnetometer at different time scales \cite{valenti2015keeping}. In optimisation, the orientation estimation is obtained based on gradient-based optimisation algorithms \cite{madgwick2011estimation, kok2014optimization}. Until recently, deep learning was introduced to estimate the orientation \cite{brossard2020denoising}, in which a deep neural network is trained to mimic the noise distribution of gyroscopes such that accurate orientation estimates can be obtained by open-loop integration of the noise-free gyro measurements. These algorithms showed superior estimation performance empirically. However, the performance can not be theoretically guaranteed, i.e., the orientation estimate error never diverges. This paper will employ Lyapunov's method in control theory to prove the estimation error boundedness guarantee using samples. Based on the theoretical result, we will develop a reinforcement learning (RL) based algorithm to learn the estimator from samples. 

RL was first applied for state estimation in \cite{morimoto2007reinforcement}. Motivated by this work, we plan to develop an RL algorithm to learn the estimator gain using samples while the orientation estimator remains the structure of conventional EKF. The key idea is, the estimator gain will be approximated by a deep neural network (DNN) as a function of the sequence of estimate errors. Unlike other popular RL algorithms \cite{lillicrap2015continuous,Schulman2017PPO,haarnoja2018SAC}, the value function will be treated as a Lyapunov function used to guarantee the estimation performance. Lyapunov's method has been widely used as a basic tool for stability analysis in control theory \cite{slotine1991applied}. To analyse the stability, the key is to find a scalar ``energy-like'' Lyapunov function for the considered system such that the derivative/difference of Lyapunov function along the state trajectory is semi-negative definite. Nonetheless, the construction/learning of the Lyapunov function is not trivial. In \cite{petridis2006construction}, a straightforward approach is proposed to construct the Lyapunov function for nonlinear systems using DNNs. Recently, the asymptotic
stability in model-free RL is given for robotic control tasks in \cite{han2020actor}. Inspired by the works \cite{petridis2006construction, han2020actor}, we will also parametrise the  Lyapunov function as a DNN and learn the parameters from samples. After that, a soft actor-critic (SAC) like algorithm \cite{haarnoja2018SAC} that incorporates the Lyapunov boundedness condition in the objective function to be optimised is proposed. Using the learned estimator gain, the estimate error of the orientation estimator is guaranteed to be bounded all the time.

In summary, we combine Lyapunov's method and DRL to design a state estimator with estimation error boundedness guarantee for orientation estimation. The main contribution of this paper has threefold:
\begin{enumerate}
    \item To the best of our knowledge, this is the first DRL-based orientation estimation method using inertial sensors combined with a magnetometer;
    \item The boundedness guarantee for estimation error is proved using Lyapunov's method in control theory;
    \item The proposed algorithm is superior for arbitrary estimation initialisation and can adapt to enormous angular velocities for which other algorithms, such as the EKF, UKF and complementary filter algorithms, can be hardly applicable.
\end{enumerate}

The rest of the paper is organised as follows. In Section \ref{sec:problem}, the orientation estimation problem is formulated. Section \ref{section: analysis}, the theoretical result on estimation error boundedness guarantee is proved. Section \ref{section: alogorithm}, a DRL algorithm based on Soft Actor-Critic (SAC) combined with theoretical results, is proposed to learn the estimator gain. In Section \ref{sec: sim and exp}, our method is compared with the EKF, UKF and complementary filter algorithms on simulated and real datasets. The conclusion is given in Section
\ref{sec: conclusion}.

\section{Problem Formulation}\label{sec:problem}
This paper uses the inertial sensors (3D accelerometers and 3D gyroscopes) combined with the magnetometer to estimate the orientation. As in \cite{kok2017using}, the system dynamics is the standard orientation dynamics in \eqref{eq:models-ssOri-dyn}. Moreover, our goal is to design the estimator gain like the classic Kalman filter. Unlike other nonlinear filtering techniques based on linearisation, we will show that the estimator gain's computation can be solved as an RL problem.
\subsection{System dynamics and state estimator}

The orientation dynamics is standard as given in~\cite{kok2017using}:
\begin{equation}\label{eq:models-ssOri-dyn}
q^\text{nb}_{t+1} = q_t^\text{nb} \odot \exp_{q} \left( \frac{T}{2} (y_{\omega,t} - e_{\omega,t} ) \right), 
\end{equation}
where $q_t^\text{nb} \in \mathbb{R}^4$ is the unit quaternion for the orientation of the body frame with respect to the navigation frame at time instant $t \in [0, T]$, $\exp_{q}(\cdot)$ corresponds to the exponential function of the quaternion, and $y_{\omega,t}$ is the gyroscope measurement. The distribution of the gyroscope noise is assumed to be Gaussian, i.e., $e_{\omega,t} \sim \mathcal{N}(0,\Sigma_\omega)$ where $\Sigma_\omega$ is the covariance matrix. 

Assuming that the linear acceleration is approximately zero, the measurement equations are given as follows: 
\begin{subequations}
\label{eq:models-ssOri}
\begin{align}
y_{\text{a},t} &= - R^\text{bn}_t g^\text{n} + e_{\text{a},t} , \label{eq:models-ssOri-measAcc} \\
y_{\text{m},t} &= R^\text{bn}_t m^\text{n} + e_{\text{m},t}, \label{eq:models-ssOri-measMag}
\end{align}
\end{subequations}
where $y_{\text{a},t}, y_{\text{m},t} \in \mathbb{R}^3$ are accelerometer and magnetometer measurements at time instant $t$ respectively, $R^\text{bn}_t$ is the rotation matrix from the navigation frame to the body frame at time instant $t$,  $g^\text{n}, m^\text{n}$ denote the local earth gravity vector and the local earth magnetic field vector, respectively. The noises $e_{\text{a},t} \sim \mathcal{N}(0, \Sigma_\text{a})$, and  $e_{\text{m},t} \sim \mathcal{N}(0,\Sigma_\text{m})$ with  $\Sigma_m = \sigma_m^2 \, \mathcal{I}_3$ and $\Sigma_\text{a} = \sigma_\text{a}^2 \, \mathcal{I}_3$. 

To estimate $q_{t+1}^\text{nb}$, the following estimator in terms of the orientation deviation is often proposed
\cite{kok2017using,crassidis2007survey}:
\begin{subequations}
\label{eq:estimator-deviation}
\begin{align}
\hat{q}^{\text{nb}}_{t+1 \mid t} &= \hat{q}^{\text{nb}}_{t \mid t} \odot \exp_{q} \left( \tfrac{T}{2} y_{\omega,t} \right),  \label{eq:oriErrorEKF_Txupdate} \\
\hat{\eta}_{t+1} &= K_{t+1} (y_{t+1}-\hat{y}_{t+1 \mid t}),   \label{eq:ori_qEkf_measUpdate} \\
\hat{q}^\text{nb}_{t+1 \mid t+1} &= \exp_{q} \left( \hat{\eta}_{t+1} \right) \odot \hat{q}^\text{nb}_{t+1 \mid t}\label{eq:ori_qEkf_comb}
\end{align}%
\end{subequations}
with 
{\small
\begin{align*}
y_{t} &= \begin{pmatrix} y_{\text{a},t} \\ y_{\text{m},t} \end{pmatrix}, 
& \hat{y}_{t+1\mid t} &=\begin{pmatrix}{-R \left\{\ \hat{q}^\text{nb}_{t \mid t} \odot \exp_{q} (\tfrac{T}{2}  y_{\omega,t})\right\}^{\top}g^n}\\{R \left\{\hat{q}^\text{nb}_{t \mid t} \odot \exp_{q} (\tfrac{T}{2}  y_{\omega,t})\right\}^{\top} m^n}\end{pmatrix},
\end{align*}
}
\hspace{-0.1cm}where $\hat{q}^\text{nb}_{t+1 \mid t}$ is the linearisation point parametrised in terms of quaternions, $\hat{\eta}_{t+1}^\text{n}$ is the state estimate of the orientation deviation, and $R\{\cdot\}$ denotes the matrix formula of translation from quaternion to rotation. The goal is to obtain $K_{t+1}$, i.e., the estimator gain at time instant $t+1$, which will be explained later in Section~\ref{sec:mdp}.

Define the orientation error 
\begin{align}\label{eq:error-def}
\tilde{q}_{t}\define q_{t}^\text{nb} \odot \left( \hat{q}^\text{nb}_{t \mid t} \right) ^\text{c},
\end{align}
or equivalently $q_{t}^\text{nb}=\tilde{q}_{t} \odot \hat{q}^\text{nb}_{t \mid t}$. From \eqref{eq:models-ssOri-dyn}, \eqref{eq:models-ssOri} and \eqref{eq:estimator-deviation}, we have
\begin{align}
&\tilde{q}_{t+1}= q_{t+1}^\text{nb}  \odot \left(  \hat{q}^\text{nb}_{t+1 \mid t+1}  \right) ^\text{c} \nonumber \\
=& \Big(\big(\tilde{q}_{t} \odot \hat{q}^\text{nb}_{t \mid t}\big) \odot \exp_{q} \big( \tfrac{T}{2} (y_{\omega,t} - e_{\omega,t} ) \big)\Big) \odot \label{eq:estimte_error} \\
&\left(\Big(\exp_{q} \big( \tfrac{1}{2} K_{t+1} \big( y_{t+1} - \hat{y}_{t+1\mid t}  \big) \big)  \odot  \hat{q}^\text{nb}_{t \mid t} \odot \exp_{q} (\tfrac{T}{2} y_{\omega,t} )\Big )  \right) ^\text{c} \nonumber
\end{align}
where, $(\cdot)^c$ denotes the conjugate of quaternion.

Furthermore, to escape the unit determinant condition of the quaternion representation of rotation, the logarithm map of the quaternion is used \cite {sola2012quaternion}:
\begin{equation}\label{eq:exp}
\begin{split}
[\eta_{t+1}]_\times = \log(\tilde{q}_{t+1})
\end{split}
\end{equation}
where $\eta_{t+1}$ is the orientation deviation and the skew operator $[\cdot]_\times $ produces the cross-product matrix. 

\subsection{Estimate Error Dynamics as Markov Decision Process}\label{sec:mdp}
By combining \eqref{eq:estimte_error} and \eqref{eq:exp}, the estimate error dynamics can actually be modelled as a Markov decision process (MDP) which
is defined as a tuple $<\mathcal{S},\mathcal{A}, \mathcal{P}, \mathcal{C},\gamma >$:
\begin{equation}\label{eq:MDP}
\tilde{q}_{t+1} \sim \mathcal{P}\left(\tilde{q}_{t+1} | \tilde{q}_{t}, K_{t+1}\right), \forall t \in \mathbb{Z}_{+},
\end{equation}
where the estimate error $\tilde{q}_{t} \in \mathcal{S}$ is the state, the estimator gain $K_{t+1} \in \mathcal{A}$ is the action sampled from a stochastic policy. Considering that the ground truth $q_{t}$ is known during training phase, the mapping between $\tilde{q}_{t}$ and $\hat{\eta}_{t}$ is bijective to some extent according to
\eqref{eq:ori_qEkf_comb} and \eqref{eq:error-def}. For convenience, in our implementation of the algorithm, we treat $\pi(K_{t+1}|\hat{\eta}_{t})$ and $\pi(K_{t+1}|\tilde{q}_{t})$ equivalently.

The state dynamics can be characterised by the transition probability function $\mathcal{P}(\tilde{q}_{t+1}|\tilde{q}_{t}, K_{t+1})$. RL algorithms can be used to find the policy $\pi$, given a cost function\footnote{We will use cost, which is often used in control literature, instead of reward. } $C(\tilde{q}_{t}, K_{t+1}) \in \mathcal{C}$ that measures the goodness of a state-action pair. In state estimation, it is often desired that the estimate error $\tilde{q}_{t}$ converges exponentially to a finite bound in mean square. As such, the cost function is selected as $C(\tilde{q}_{t}, K_{t+1})=\mathbb{E}_{P(\cdot|\tilde{q}_{t},K_{t+1})}[\|\tilde{q}_{t+1}\|^2]$, and the return is the sum of discounted cost $\sum_{\tau=t}^{\infty}\gamma^{\tau-t}C(\tilde{q}_{t}, K_{t+1})$ with the discount factor $\gamma\in [0,~1)$, where $\mathbb{E}[\cdot]$ denotes the expected value. 

\begin{definition}{\cite{reif1999stochastic} }
The estimate error $\tilde{q}_{t}$ in the MDP \eqref{eq:MDP} is said to be exponentially bounded in mean square if $\exists~\eta>0$ and $0<\varphi<1$, such that
\begin{equation}\label{covergence}
\vspace{-0.2cm}
\mathbb{E}[\|\tilde{q}_{t}\|^2]\leq \eta\mathbb{E}{[\|\tilde{q}_{0}\|^2]}\varphi^{t}+p, 
\vspace{-0.1cm}
\end{equation}
holds at all the time instants $t\geq 0$, where $p$ is a positive constant number.
\end{definition}

In this paper, our goal is to learn the estimator gain $K_{t+1} =\pi(\hat{\eta}_{t})$ in  \eqref{eq:estimator-deviation} which can be seen as a policy obtained using an RL algorithm, such that the mean square of the estimate error of $\tilde{q}_{t}$ in \eqref{eq:MDP} is guaranteed to converge exponentially to a positive bound. Different from the EKF where $K_{t+1}$ is computed using the linearisation approximation, in this paper $K_{t+1}$ is approximated by a DNN $\pi(\cdot)$. 

\vspace{-0.1cm}
\section{Estimation Error Boundedness Guarantee}\label{section: analysis}
\vspace{-0.1cm}
In this section, we propose the main theorem to guarantee the boundedness of the estimate error. Before proceeding, some notations need be clarified. $\rho(\tilde{q}_0)$ denotes the distribution of the starting state estimate error $\tilde{q}_0$. The state distribution of state estimate error at a certain instant $t$ as $P(\tilde{q}_{t}|\rho,\pi,t)$ is defined in an iterative way: $P(\tilde{q}_{t+1}=s'|\rho,\pi,t+1)=\int_{S}P(\tilde{q}_{t}=s|\rho,\pi,t)P_{\pi}(s'|s)ds$. The following assumption, which is often used in RL literature, is needed:
\begin{assumption}
The Markov chain in \eqref{eq:MDP} induced by a policy $\pi$ is ergodic with a unique distribution probability. That is,
$\exists\ p_{\pi}(s),$ such that
\begin{equation}\label{eq:ergodic}
    p_{\pi}(s)=\lim\limits_{t\rightarrow \infty}P(\tilde{q}_t=s|\rho,\pi,t)
\end{equation}
\end{assumption}
\begin{theorem}\label{theorem}
The error dynamics \eqref{eq:MDP} is exponentially bounded in mean square if there exists a Lyapunov function $L(\tilde{q}_t):S\rightarrow R^{+}$ and positive constants $\alpha_{1}, \alpha_{2}$ and $\delta$
such that 
\begin{equation}\label{uplowerbound}
    \alpha_{1}\mathbb{E}_{ \pi}[\|\tilde{q}_{t}\|^2]-\delta\leq L(\tilde{q}_{t}) \leq \alpha_{1}\mathbb{E}_{ \pi}[\|\tilde{q}_{t}\|^2]
\end{equation}
and
\begin{equation}\label{inequality}
\begin{split}
    \lim\limits_{N\rightarrow +\infty}[\ln(\mathbb{E}_{\tilde{q}_{t}\sim \mu_{N}}(\mathbb{E}_{\tilde{q}_{t+1}\sim P_{\pi}}L(\tilde{q}_{t+1})))\\
    -\mathbb{E}_{\tilde{q}_{t}\sim \mu_{N}}\ln(L(\tilde{q}_{t}))]\leq -\alpha_{2}
\end{split}
\end{equation}
where 
\begin{equation}
    \mu_{N}(s)\define  \frac{1}{N}\sum\limits_{t=0}^{N-1}P(\tilde{q}_{t}=s|\rho,\pi,t)
\end{equation}
\end{theorem}
Proof: We have
\begin{equation}
    \begin{split}
        &\ln(\mathbb{E}_{\tilde{q}_{t}\sim \mu_{N}}(\mathbb{E}_{\tilde{q}_{t+1}\sim P_{\pi}}L(\tilde{q}_{t+1}))\\
        =&\ln(\int_{S}\frac{1}{N}\sum\limits_{t=0}^{N-1}P(\tilde{q}_{t}=s|\rho,\pi,t)\int_{S}P_{\pi}(s'|s)L(s')\,ds'\,ds)
        \\=&\ln(\int_{S}(\int_{S}\frac{1}{N}\sum\limits_{t=0}^{N-1}P(\tilde{q}_{t}=s|\rho,\pi,t)P_{\pi}(s'|s)\,ds)L(s')\,ds')
        \\=&\ln(\int_{S}(\frac{1}{N}\sum\limits_{t=0}^{N-1}P(\tilde{q}_{t+1}=s'|\rho,\pi,t+1))L(s')\,ds')
        \\=&\ln((\frac{1}{N}\sum\limits_{t=0}^{N-1}\int_{S}P(\tilde{q}_{t+1}=s'|\rho,\pi,t+1))L(s')\,ds')
        \\\geq& \frac{1}{N}\sum\limits_{t=0}^{N-1}\ln((\int_{S}P(\tilde{q}_{t+1}=s'|\rho,\pi,t+1))L(s')\,ds')
    \end{split}
\end{equation}
where the last inequality follows from the fact that  $\ln(x)$ is a concave function on $R^{+}$. Similarly, noting that $-\ln(x)$ is a convex function we have
\begin{equation}
    \begin{split}
        &-\mathbb{E}_{\tilde{q}_{t}\sim \mu_{N}}\ln L(\tilde{q}_{t})
        \\=&-\int_{S}\frac{1}{N}\sum\limits_{t=0}^{N-1}P(\tilde{q}_{t}=s|\rho,\pi,t)\ln(L(s))\,ds
        \\=&\frac{1}{N}\sum\limits_{t=0}^{N-1}\int_{S}P(\tilde{q}_{t}=s|\rho,\pi,t)(-\ln L(s))\,ds
        \\\geq& \frac{1}{N}\sum\limits_{t=0}^{N-1}-\ln(\int_{S}P(\tilde{q}_{t}=s|\rho,\pi,t)L(s)\,ds)
    \end{split}
\end{equation}
It follows from the above two inequalities that 
\begin{equation}
    \begin{split}
        &\ln(\mathbb{E}_{\tilde{q}_{t}\sim \mu_{N}}(\mathbb{E}_{\tilde{q}_{t+1}\sim P_{\pi}}L(\tilde{q}_{t+1}))-\mathbb{E}_{\tilde{q}_{t}\sim \mu_{N}}\ln L(\tilde{q}_{t}))
        \\ \geq& \frac{1}{N}\sum\limits_{t=0}^{N-1}\ln\frac{\int_{S}P(\tilde{q}_{t+1}=s'|\rho,\pi,t+1)L(s')\,ds'}{\int_{S}P(\tilde{q}_{t}=s|\rho,\pi,t)L(s)\,ds}
        \\\geq&\frac{1}{N}\sum\limits_{t=0}^{N-1}\ln\frac{\mathbb{E}_{\tilde{q}_{t+1}}L(\tilde{q}_{t+1})}{\mathbb{E}_{\tilde{q}_{t}}L(\tilde{q}_{t})}
    \end{split}
\end{equation}
Substituting the above into \eqref{inequality},  we obtain 
\begin{equation}
    \lim\limits_{N\rightarrow +\infty}\frac{1}{N}\sum\limits_{t=0}^{N-1}\ln\frac{\mathbb{E}_{\tilde{q}_{t+1}}L(\tilde{q}_{t+1})}{\mathbb{E}_{\tilde{q}_{t}}L(\tilde{q}_{t})}\leq -\alpha_2
\end{equation}
then 
\begin{equation}
    \lim\limits_{N\rightarrow +\infty}\frac{1}{N}\ln\frac{\mathbb{E}_{\tilde{q}_{N}}L(\tilde{q}_{N})}{\mathbb{E}_{\tilde{q}_{0}}L(\tilde{q}_{0})} \leq -\alpha_2
\end{equation}
It means that $\forall \epsilon>0, \exists N_{\epsilon}$, $\frac{1}{N}\ln\frac{\mathbb{E}_{\tilde{q}_{N}}L(\tilde{q}_{N})}{\mathbb{E}_{\tilde{q}_{0}}L(\tilde{q}_{0})}<-\alpha_2 + \epsilon<0$ holds when $ N>N_{\epsilon},$ namely
\begin{equation}\label{exp_stable}
    \frac{\mathbb{E}_{\tilde{q}_{N}}L(\tilde{q}_{N})}{\mathbb{E}_{\tilde{q}_{0}}L(\tilde{q}_{0})}\leq\mathrm{e}^{N(-\alpha_{2}+\epsilon)},\forall N>N_{\epsilon}
\end{equation}
So we get for sufficiently large $N>N_{\epsilon}$,
\begin{equation}
    \mathbb{E}_{\tilde{q}_{N}\sim P(\tilde{q}_{N}|\rho,\pi,N)}L(\tilde{q}_{N})\leq \mathrm{e}^{N(-\alpha_2+\epsilon)}\mathbb{E}_{\tilde{q}_{0}\sim \rho(\tilde{q}_{0})}L(\tilde{q}_{0})
\end{equation}
By Equation \eqref{uplowerbound} we have the following result
\begin{equation}
\begin{split}
    &\mathbb{E}_{\tilde{q}_{N}\sim P(\tilde{q}_{N}|\rho,\pi,N)}\mathbb{E}_{\pi}\|\tilde{q}_N\|^2\\
    \leq& \mathrm{e}^{N(-\alpha_2+\epsilon)}\mathbb{E}_{\tilde{q}_{0}\sim \rho(\tilde{q}_{0})}\mathbb{E}_{\pi}\|\tilde{q}_0\|^2+\frac{\delta}{\alpha_1}
\end{split}
\end{equation}

So far, it has been proved that the estimate error $\tilde{q}_t$ in \eqref{eq:MDP} is exponentially bounded according to Definition 1.

\section{Lyapunov-based Reinforcement Learning Orientation Estimation Algorithm}\label{section: alogorithm}

\begin{figure}[tb]
    \centering
    \includegraphics[width=0.39\textwidth]{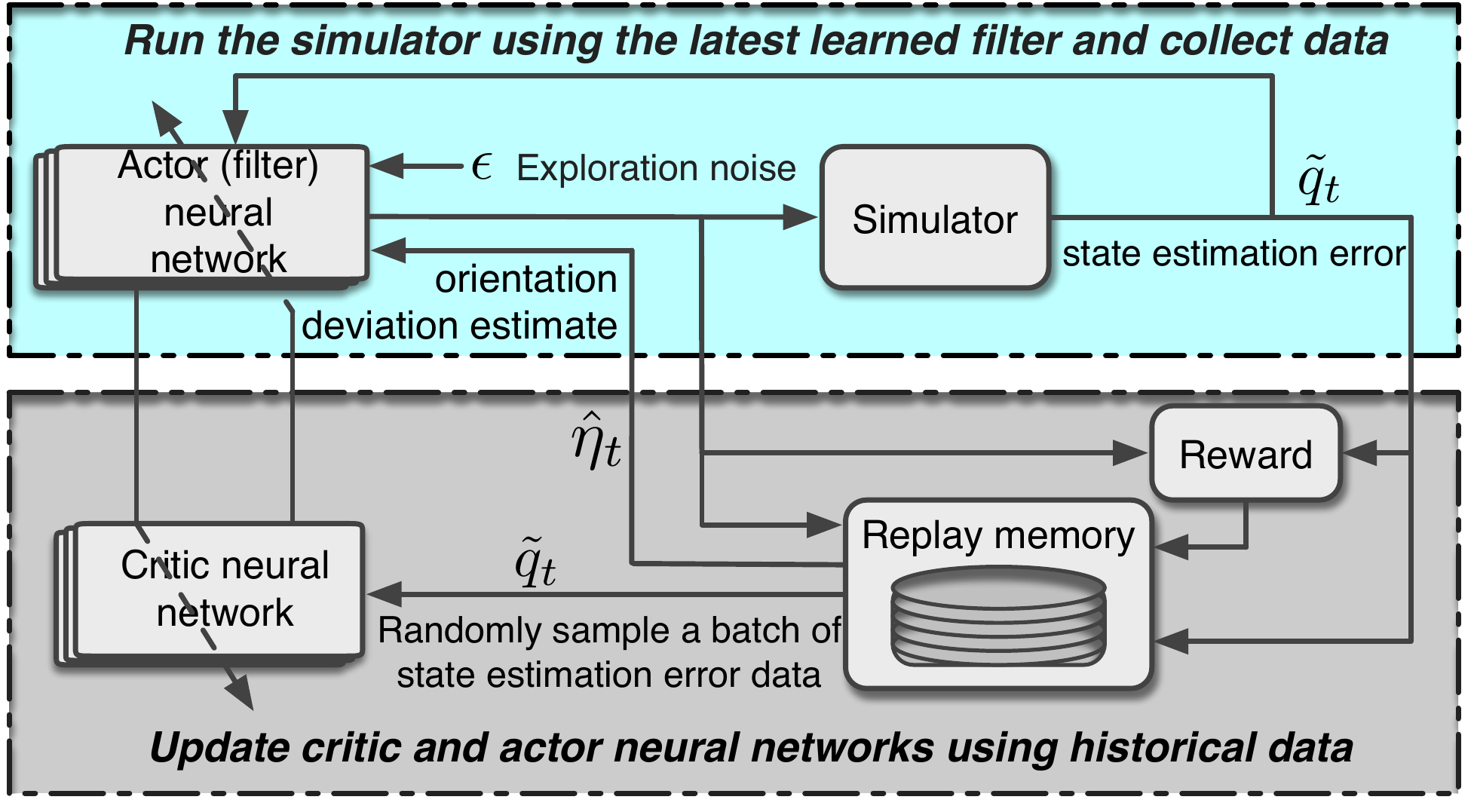}
    \vspace{-0.2cm}
    \caption{Offline RL training process of orientation state estimator}
    \vspace{-0.3cm}
    \label{fig:TrainRL}
\end{figure}

In this section, we will combine SAC algorithm \cite{haarnoja2018SAC}, one of the state-of-the-art RL algorithms, with the theoretical result in Section~\ref{section: analysis} to learn the gain/policy $K_{t+1}$ for the state estimator \eqref{eq:estimator-deviation}.

Considering MDP in \eqref{eq:MDP}, the orientation estimation problem can be viewed as a RL problem in which the policy is sought after by minimising the expected accumulated cost. Here a stochastic policy is chosen as  $\pi(K_{t+1}\mid\tilde{q}_{t}) \sim \mathcal{N}(K_{t+1}(\tilde{q}_{t}),\sigma)$ from which the gain $K_{t+1}$ for a given state $\tilde{q}_{t}$ is sampled \cite{Sutton2018MIT}. The corresponding Q-function (a.k.a, state-action value function) is given as:
\begin{equation}\label{eq:Q func}
 Q_\pi(\tilde{q}_{t},K_{t+1})=C_t(\tilde{q}_{t},K_{t+1})+\gamma \mathbb{E}_{\tilde{q}_{t+1}}[V_{\pi}(\tilde{q}_{t+1})]    
\end{equation}
To this end, $K_{t+1}$ can be learned by many existing RL algorithms.

Motivated by the works in \cite{han2020actor,Hu2020Lyapunov,zhang2020model1,zhang2020model2}, we propose to incorporate the theoretical result in Theorem \ref{theorem} to policy optimisation in SAC as a constrained optimisation problem. First of all, a Lyapunov candidate needs to be selected at the first instance. In the context of RL, a Lyapunov candidate will be parametrised/selected as the Q-function \cite{petridis2006construction,perkins2002lyapunov}. In this paper, we choose $L(\tilde{q}_t)$ in \eqref{uplowerbound} as: 
\begin{equation}\label{Lyapunov}
   L_(\tilde{q}_t)=\mathbb{E}_{K_{t+1}\sim \pi}[L_c(\tilde{q}_t,K_{t+1})]
\end{equation}
where $L_c(\tilde{q}_t,K_{t+1}) =Q(\tilde{q}_t,K_{t+1})$. 
Then the constrained optimisation problem is:
\begin{equation}\label{optimised-pro}
\begin{aligned}
\min_{\pi}& \quad Q_\pi(\tilde{q}_{t},K_{t+1})\\
\text{s.t.}& \quad \eqref{uplowerbound} \text{ and } \eqref{inequality} \\
&  -\ln( \pi(K_{t+1}\mid\tilde{q}_{t})) \geq \mathcal{H}_t
\end{aligned}
\end{equation}
where $Q_\pi(\tilde{q}_{t},K_{t+1})$ is defined in \eqref{eq:Q func}, the second constraint is the minimum entropy constraint used in the SAC to improve the exploration in the action space \cite{Haarnoja2018SAC1} where $\mathcal{H}_t$ is the desired bound.

Denote the parametrised actor and critic as $\pi_{\theta}(K_{t+1}|\tilde{q}_{t})$ and $Q_\phi(\tilde{q}_t,K_{t+1})$ respectively, where $\theta$ and $\phi$ are the parameters of the DNNs. To ensure the positiveness of a Lyapunov function, $L_\phi(\tilde{q}_t,K_{t+1})$ is selected as the square of a DNN as $L_\phi(\tilde{q}_t,K_{t+1})=f^{\top}_\phi(\tilde{q}_t,K_{t+1})f_\phi(\tilde{q}_t,K_{t+1})$, where $f$ is the vector output of a DNN parameterised by $\phi$. On the other hand, the stochastic policy $\pi_{\theta}(K_{t+1}|\tilde{q}_{t})$ is parametrised by a DNN $f_{\theta}$ that depends on the state $\tilde{q}_{t}$ and a Gaussian noise $\epsilon$.

Solving the above constrained optimisation problem is equivalent to minimising the following objective function:
\begin{equation}
\begin{small}
\begin{aligned}
J(\theta)&=\mathbb{E}_{\tilde{q}_t,a_t,\tilde{q}_{t+1},c_{t}\sim D}[\alpha(\ln(\pi_{\theta}(f_{\theta}(\tilde{q}_t,\epsilon)|\tilde{q}_t))+\mathcal{H}_{t})\\
    &+\lambda\big(\ln{L_{\phi}(\tilde{q}_{t+1},f_{\theta}(\tilde{q}_{t+1},\epsilon))}-\ln{L(\tilde{q}_t,a_t)}+\alpha_2\big)]
\label{eq:objective}
\end{aligned}
\end{small}
\end{equation}
where $\mathcal{D}$ is the replay memory of the training samples, $\alpha$ and $\lambda$ are Lagrange multipliers which control the relative importance of constraints in \eqref{optimised-pro}.

In the actor-critic framework, the policy network parameters are updated through stochastic gradient descent of \eqref{eq:objective}. The training process can be seen in Fig.~\ref{fig:TrainRL}. It can be proved that the policy can converge to an optimal one that ensures the orientation estimate error $\mathbb{E}[\|\tilde{q}_t\|^2]$ converges to a constant as $t\to \infty, \forall  \tilde{q}_t \in S$. The proof is standard and omitted due to page limits. The readers can refer to Section IV-D in \cite{Hu2020Lyapunov} for more details. The pseudocode of the proposed Lyapunov-based reinforcement learning orientation estimation (LRLOE) algorithm is showed in Algorithm~\ref{alg1}.

\begin{algorithm}
\caption{LRLOE algorithm}
\label{alg1}
\begin{algorithmic}[1]

\State Set the initial parameters $\phi$  for the Lyapunov function $\mathcal{L}_{\phi}$, $\theta$ for the estimator gain policy $\pi_{\theta}(K_{1}|\tilde{q}_{0})$, $\lambda$ for the Lagrangian multiplier, $\alpha$ for the temperature parameter, and the replay memory $\mathcal{D}$
 
\State Set the target parameter $\bar{\theta}$ as $\bar{\theta}\leftarrow \theta$

\While{Training}
    \For {each data collection step}
        \State Choose estimator gain $K_{k+1}$ using $\pi_{\theta}(K_{k+1}|\tilde{q}_{k})$ 
        \State Simulate \eqref{eq:models-ssOri-dyn} and \eqref{eq:models-ssOri} with the orientation estimator \eqref{eq:estimator-deviation} to collect samples $\tilde{q}_{k}$
        \State $\mathcal{D} \leftarrow \mathcal{D} \cup \tilde{q}_{k}$
    \EndFor
    \State update $L_\phi$, $\pi_\theta$, $\lambda$, $\alpha$ by optimising \eqref{eq:objective}
\EndWhile
\State Output $\theta^{\ast}$, $\phi^{\ast}$, $\lambda^{\ast}$, and $\alpha^{\ast}$ 
\end{algorithmic}
\end{algorithm}

\section{Experimental Results}\label{sec: sim and exp}
In this Section, we train and infer on both simulated and real datasets. The estimation results has exhibited good performance, compared with three well-known orientation estimation algorithms: EKF \cite{marins2001extended}, UKF \cite{kraft2003quaternion}, and complementary filter \cite{valenti2015keeping}.

\subsection{Results for simulated dataset}
The RL policy is trained on a relatively trivial profile (see figure.~\ref{fig:traindata1}), then tested on three other independent profiles (see figure.~\ref{fig:inferdata1}, \ref{fig:inferdata2} and~\ref{fig:inferdata3}). The sampling rate is set to 100 Hz (consistent with real data in Section \ref{sec:realdata}). The sensor noise are sampled with the following distribution \cite{kok2017using}:
\begin{equation}
\begin{split}
    e_{\omega,t} \sim \mathcal{N}(0,\Sigma_\omega),\qquad \Sigma_\omega=0.0003, \\
    e_{\text{a},t} \sim \mathcal{N}(0, \Sigma_\text{a}),\qquad \Sigma_\text{a}=0.0005, \\
    e_{\text{m},t} \sim \mathcal{N}(0,\Sigma_\text{m}),\qquad \Sigma_\text{m}=0.0003
\end{split}
\end{equation}

{\small
\begin{table}
\caption{Hyperparameters of the proposed estimator}
\label{tab:Hp}
\begin{center}
\small
\begin{tabular}{lcccc}
    \toprule
    {Hyperparameters} & Value\\
    \midrule
    Time horizon & 1000 \\
    SGD batch size & 256 \\
    Actor learning rate & 1e-4\\
    Critic learning rate & 3e-4\\
    Lyapunov learning rate & 3e-4\\
    Soft replacement ($\tau$) & 5e-3\\
    Discount ($\gamma$) & 0.999\\
    Structure of $a_{\phi}$ & (128,64,32) \\
    Structure of $L_{\theta}$ & (128,64,32) \\
    \bottomrule
    \vspace{-1cm}
\end{tabular}
\normalsize
\end{center}
\end{table}
}

During inference, the covariance of the measurement noise $\Sigma_\omega$ is increased to $0.03$. The hyperparameters of Algorithm 1 are shown in Table \ref{tab:Hp}. The last $20\%$ of training data is used for validation. We independently train 20 policies and select the one with the lowest validation error for inference. The initial state estimate is sampled from a normal distribution with actual initial orientation as mean and a standard deviation of 0.1. The inference results of our proposed method are shown in Fig.~\ref {fig:exp cov=0.0003 initial estimation=0.1}, for all three different angular velocity profiles with accurate estimation. 

{\small
\begin{figure}
\centering
\subfigure[Training profile]{
\label{fig:traindata1}
\includegraphics[scale=0.23]{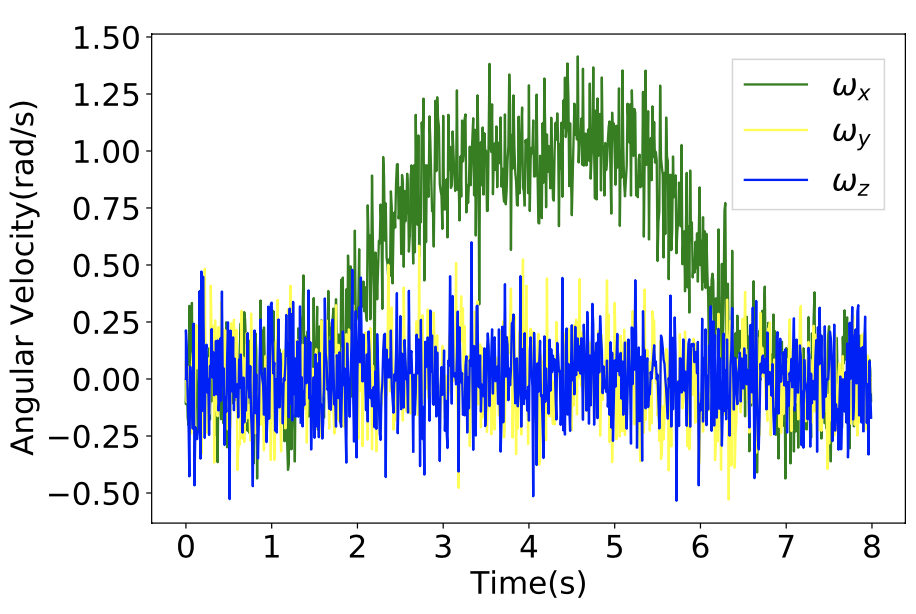}}
\subfigure[Inference profile (simple)]{
\label{fig:inferdata1}
\includegraphics[scale=0.23]{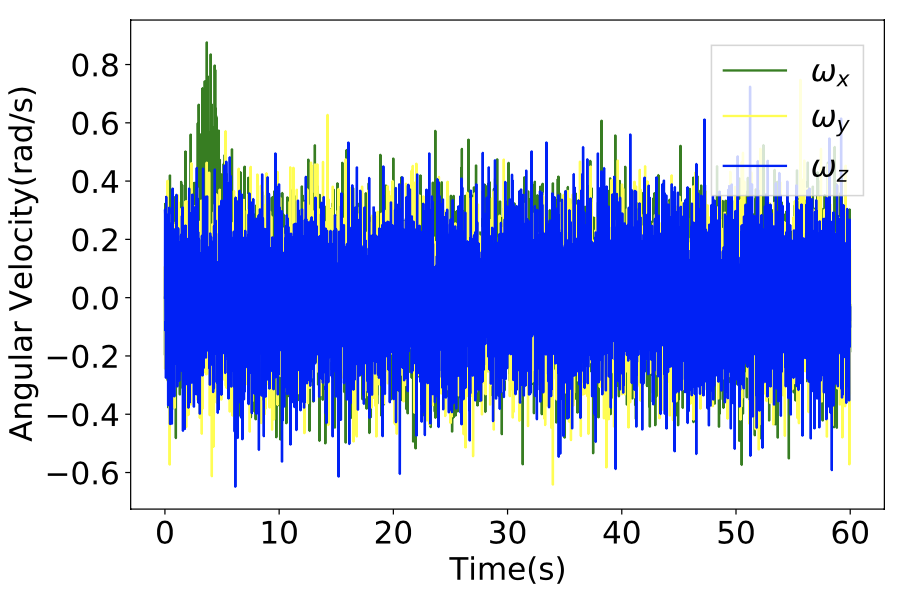}}

\vspace{-0.3cm}
\subfigure[Inference profile (medium)]{
\label{fig:inferdata2}
\includegraphics[scale=0.23]{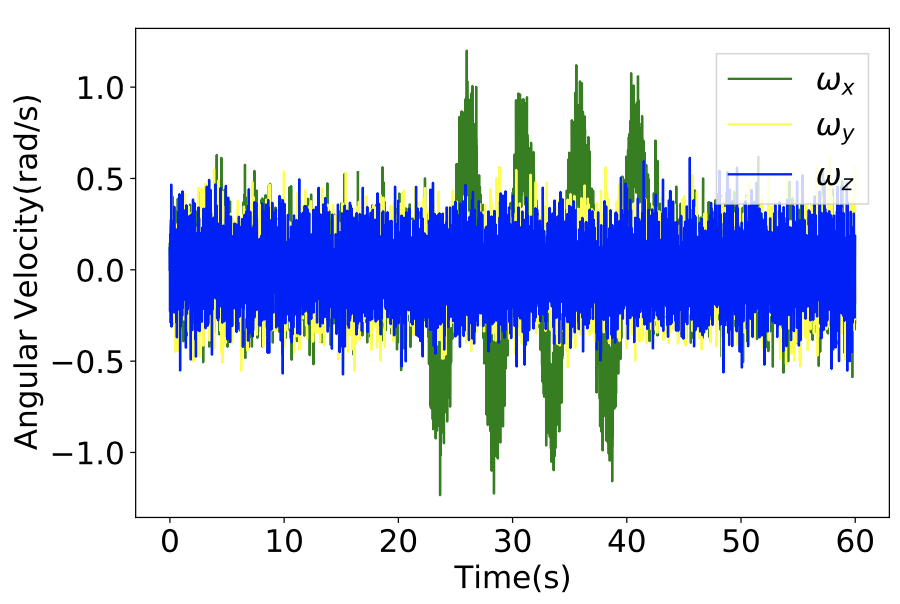}}
\subfigure[Inference profile (complicated)]{
\label{fig:inferdata3}
\includegraphics[scale=0.23]{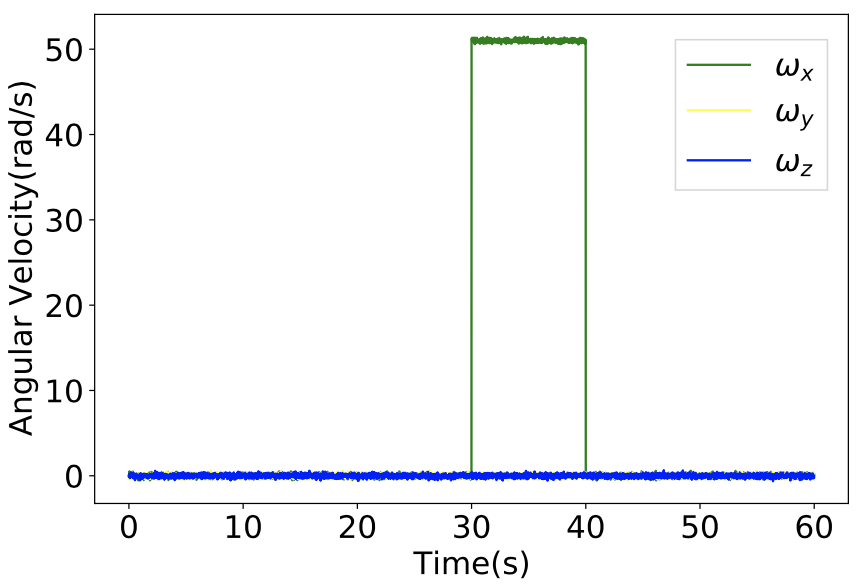}}
\vspace{-0.3cm}
\caption{The angular velocity profiles used for training and inference.
}
\vspace{-0.5cm}
\label{fig:trainingProfile}
\end{figure}

}

\begin{figure}[h]
\centering
\includegraphics[scale=0.22]{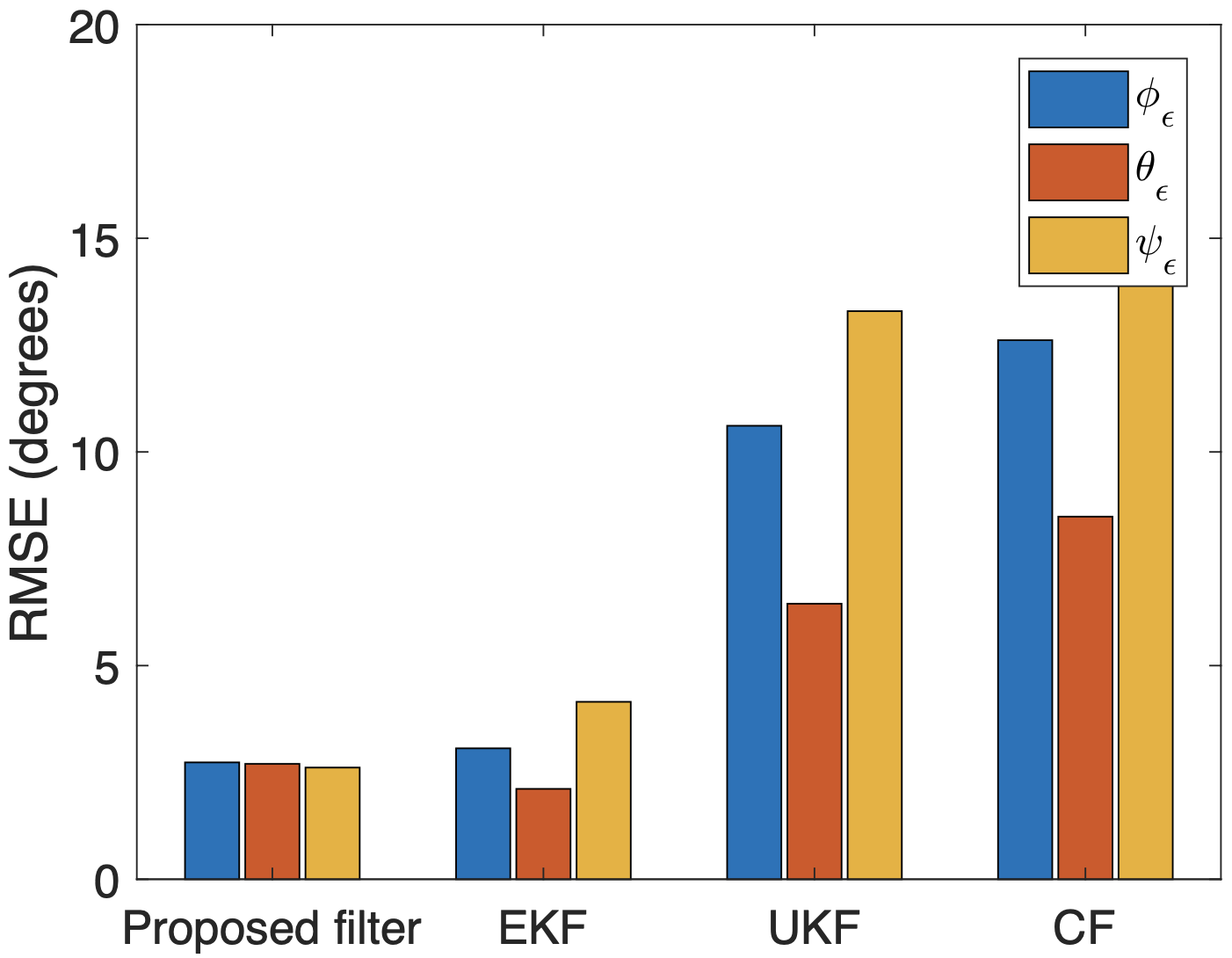}
\vspace{-0.3cm}
\caption{The RMSE of the proposed filter, Extended Kalman filter (EKF), Unscented Kalman filter (UKF) and Complementary filter (CF) for the inference profiles in Fig.~\ref{fig:inferdata2}, ~\ref{fig:inferdata2} and ~\ref{fig:inferdata3}. Because the quaternions double-cover the rotation space which imposes ambiguity in comparison, we use Euler angles instead of quaternions here.} 
\label{fig:RMSE}
\end{figure}

Furthermore, we compare our proposed DRL-based estimation method with the EKF \cite{marins2001extended}, the UKF \cite{kraft2003quaternion}, and the complementary filter (CF) \cite{valenti2015keeping}. The root of mean square errors (RMSE) of the decoupled Euler parameters for pitch, $\phi$, roll, $\theta$ and heading, $\psi$ angles, corresponding to rotations around x, y, z-axis respectively, is chosen as the estimation performance and $200$ Monte Carlo simulations are run for each algorithm. As shown in Fig.~\ref{fig:RMSE}, the DRL-based estimation method can achieve good performance using the inference profiles in Fig.~\ref{fig:inferdata2}.

\begin{figure}[h!]
\vspace{-0.1cm}
\centering
\subfigure[EKF based quaternion estimation for Fig.\ref{fig:inferdata3}]{
\vspace{-0.4cm}
\label{fig:complicated_EKF}
\includegraphics[scale=0.26]{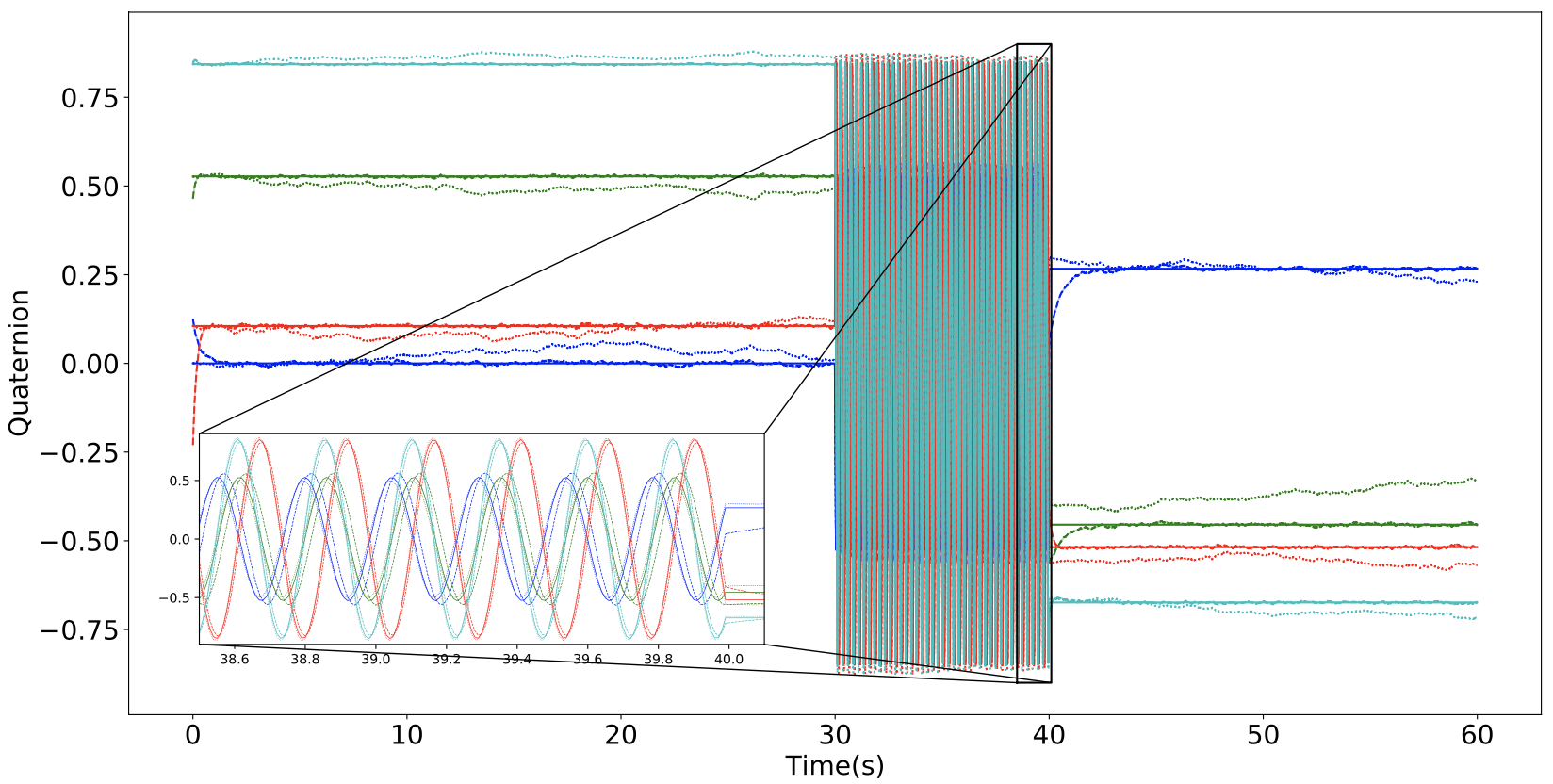}}
\subfigure[UKF based quaternion estimation for Fig.\ref{fig:inferdata3}]{
\vspace{-0.4cm}
\label{fig::complicated_UKF}
\includegraphics[scale=0.26]{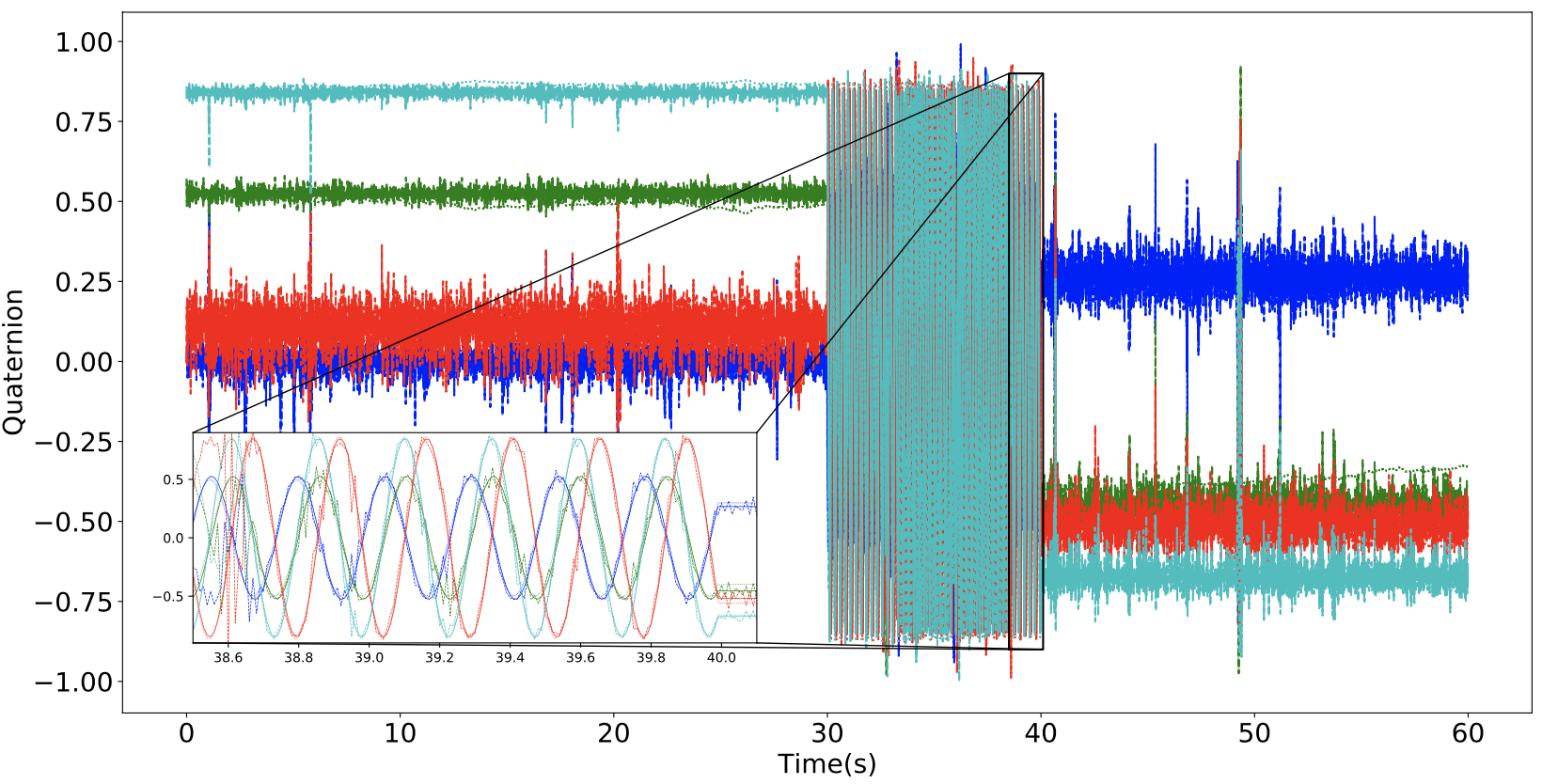}}
\subfigure[CF based quaternion estimation in Fig.\ref{fig:inferdata3}]{
\vspace{-0.4cm}
\label{fig::complicated_CF}
\includegraphics[scale=0.26]{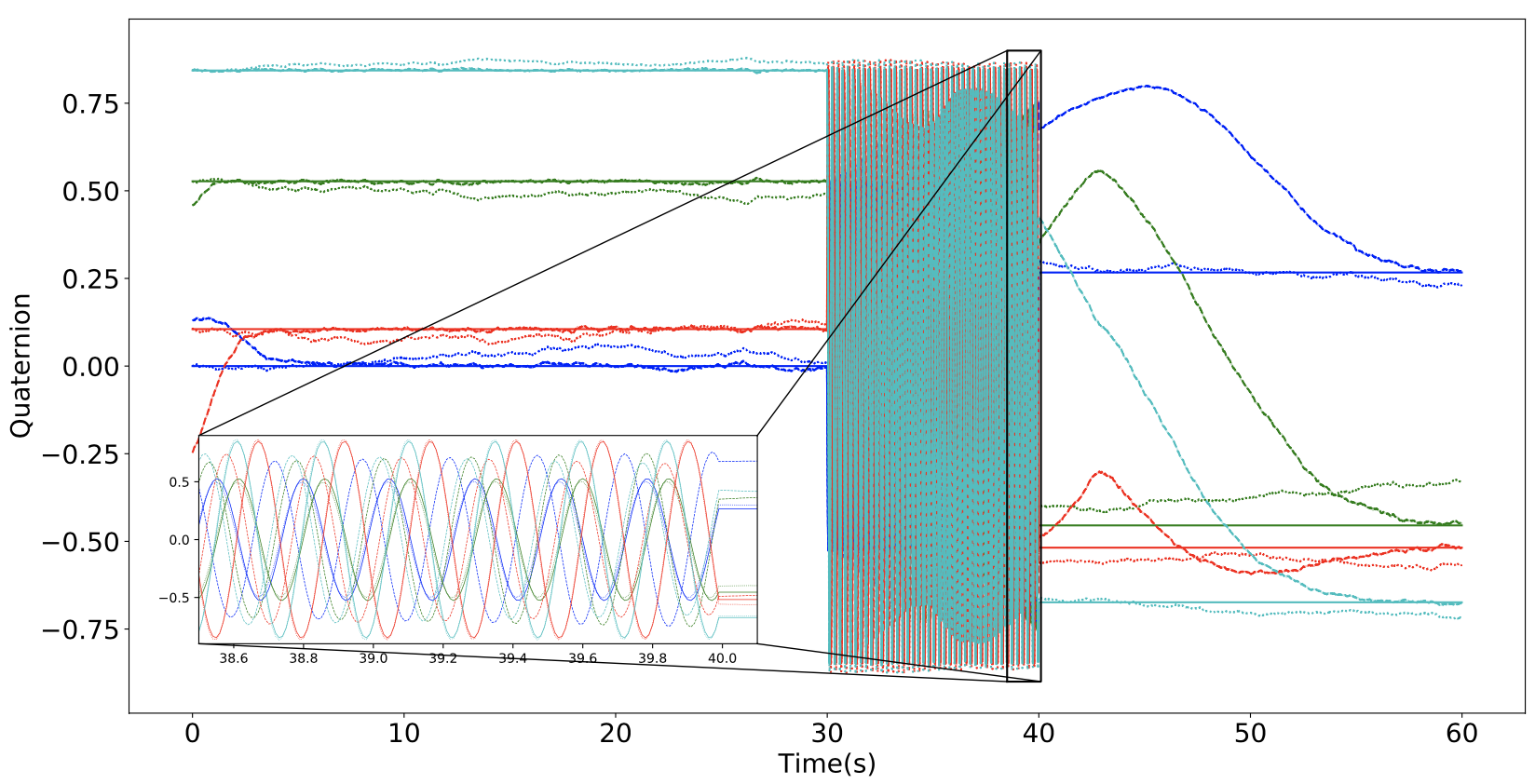}}
\centering 
\vspace{-0.1cm}
\caption{
Comparison with other classic estimation algorithms for a drastic angular velocity profile in Fig.~\ref{fig:inferdata3}. 
}
\label{fig:complicated_kalman}
\end{figure}

For a drastic angular velocity profile, the orientation estimation results of the EKF, UKF and CF  can be challenging, as shown in Fig.~\ref{fig:complicated_kalman}. It can be found that the UKF yields a significant estimate error under a high noise level, and both the EKF and complementary filter perform poorly as the estimate error accumulates in the long period of rapid rotation. In comparison, our proposed DRL-based estimation method is more robust for "wild" movement profiles.

\begin{figure}[h]
\centering
\subfigure[Quaternion for Fig.~\ref{fig:inferdata1}]{
\vspace{-0.4cm}
\label{fig:inferq1}
\includegraphics[scale=0.21]{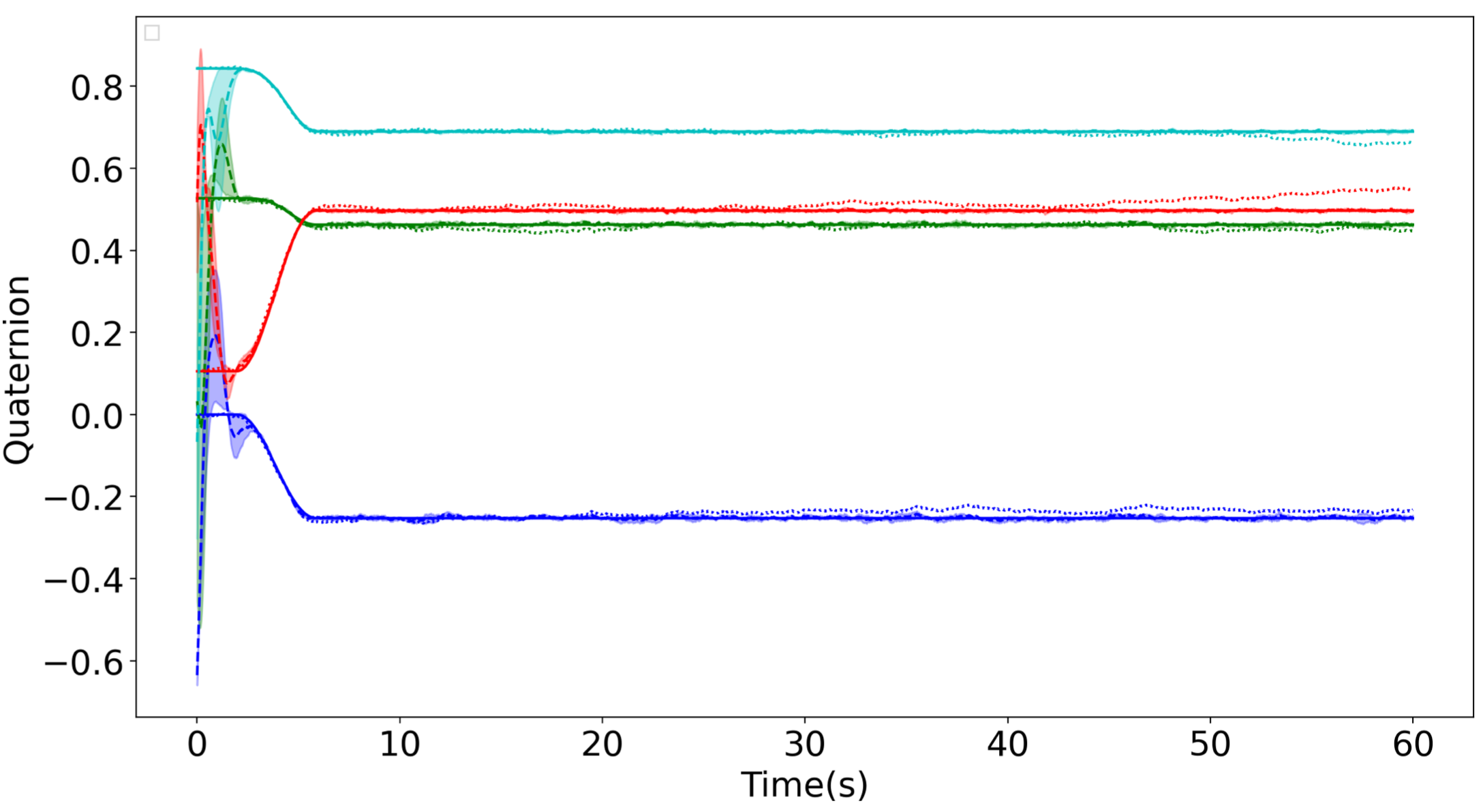}}
\subfigure[Quaternion for Fig.~\ref{fig:inferdata2}]{
\vspace{-0.4cm}
\label{fig:inferq2}
\includegraphics[scale=0.21]{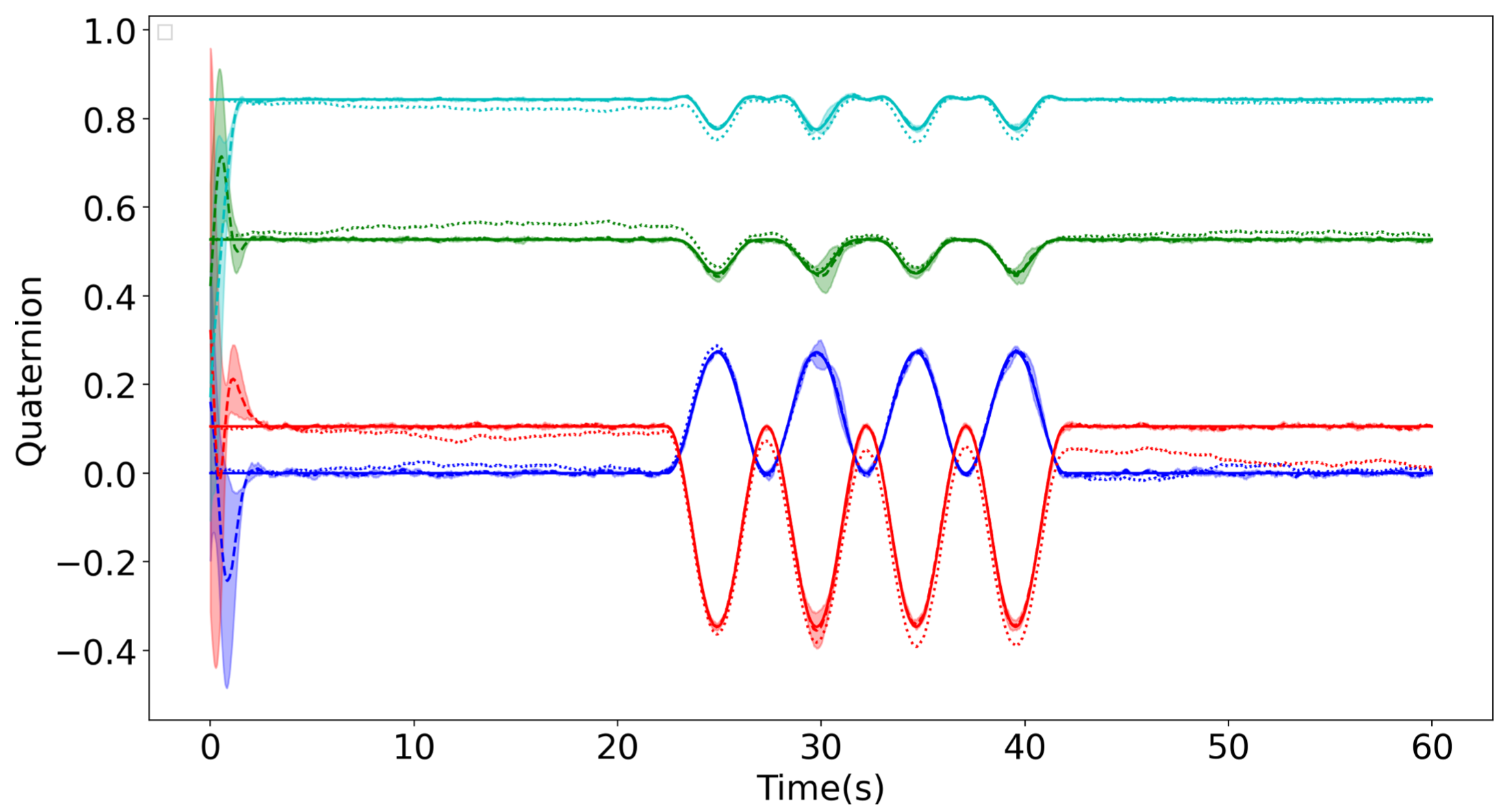}}
\subfigure[Quaternion for Fig.~\ref{fig:inferdata3}]{
\vspace{-0.4cm}
\label{fig:inferq3}
\includegraphics[scale=0.21]{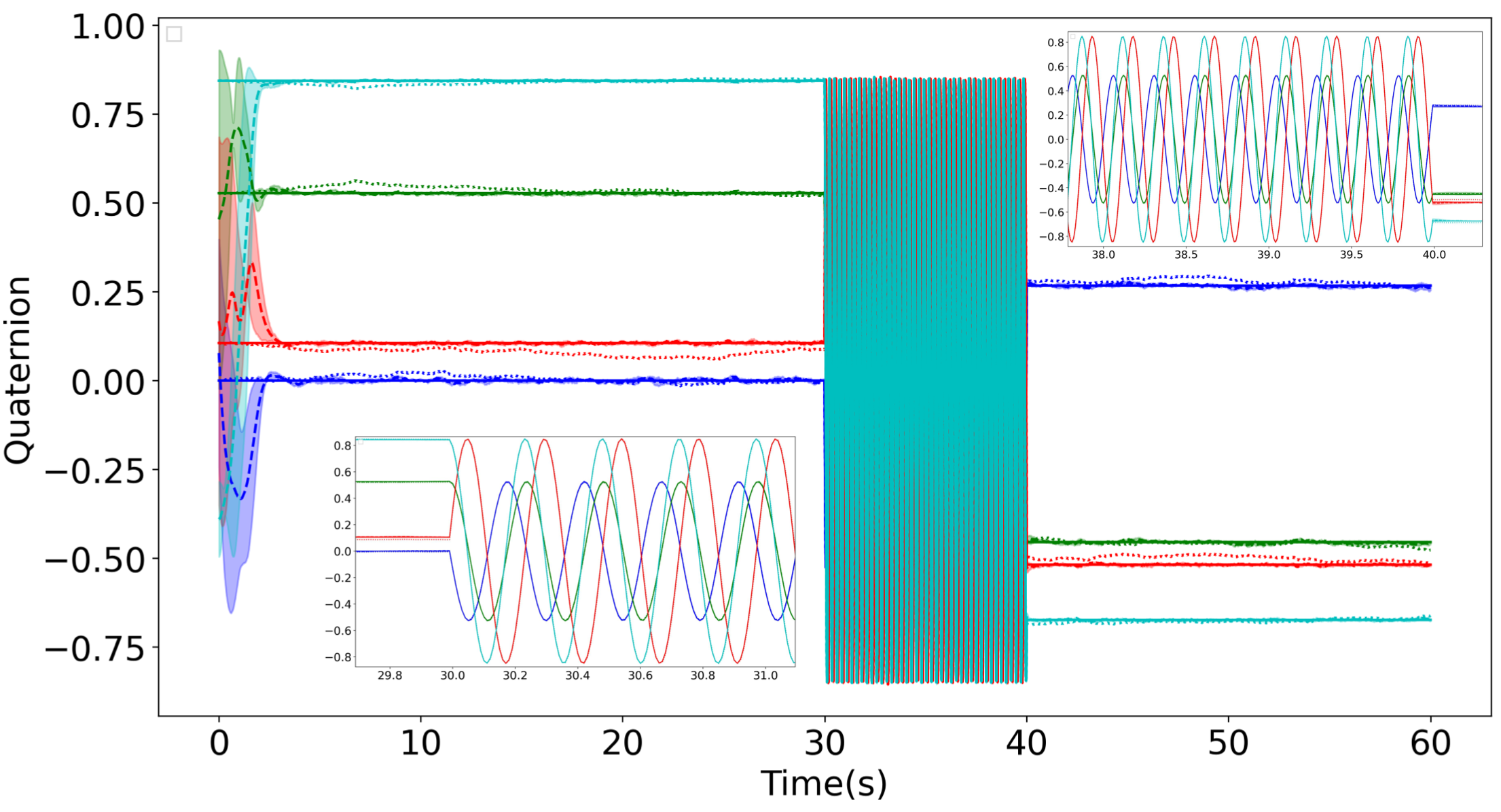}}
\caption{
Quaternion for different angular velocities in Fig.~\ref{fig:inferdata1},~\ref{fig:inferdata2} and~\ref{fig:inferdata3}. The solid, dotted and dashed lines correspond to ground truth, integrated and estimated quaternion, respectively. The shaded areas correspond to the standard deviation of over $20$ independent runs. 
For the high-velocity profile, a zoom-in for the high angular velocity period is also showed in Fig.~\ref{fig:inferq3}.
}
\label{fig:exp cov=0.0003 initial estimation=0.1}
\end{figure}

\vspace{-0.1cm}

\subsection{Results for real dataset}
\label{sec:realdata}
\begin{figure}
\centering
\includegraphics[scale = 0.3]{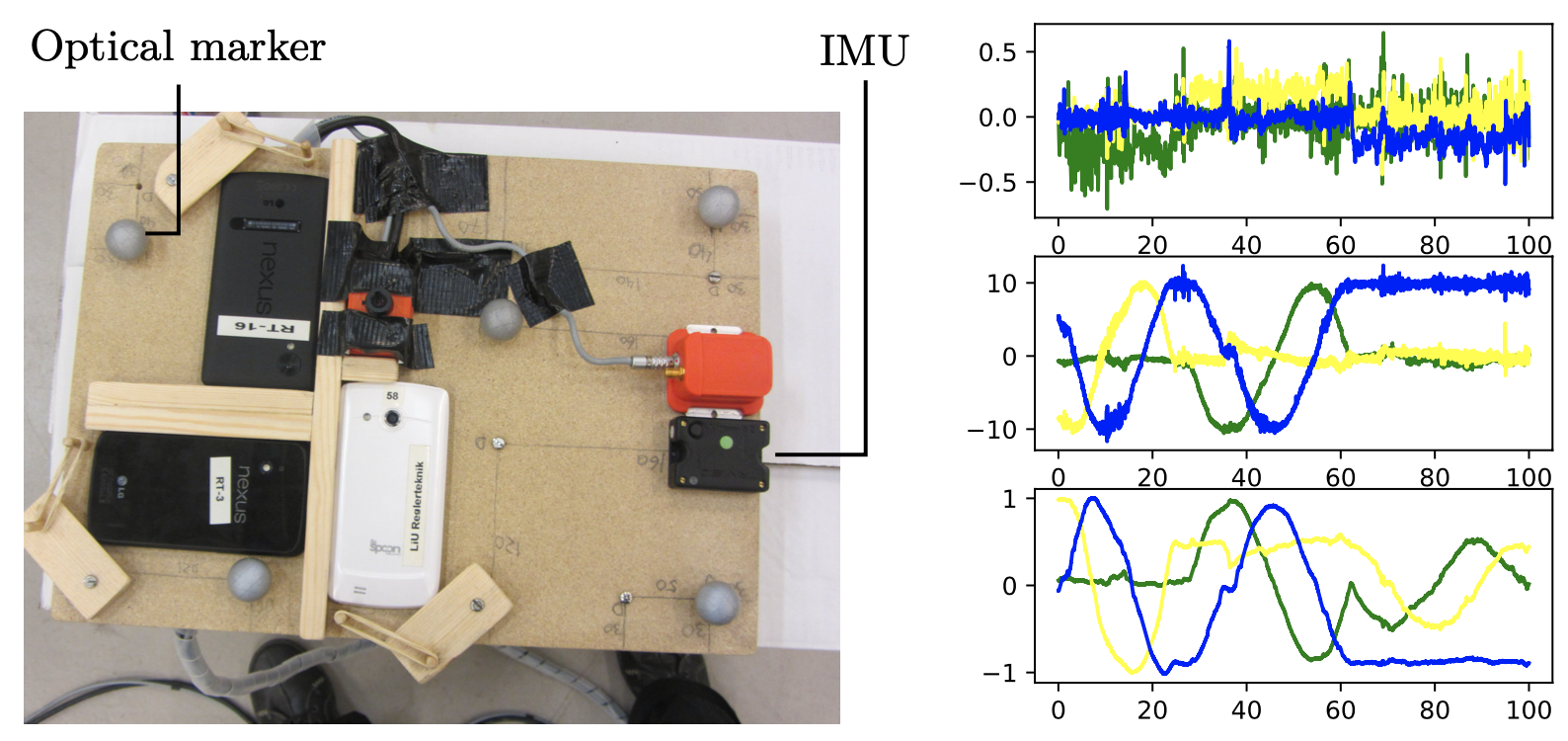}
  \caption{Real dataset (adapted from Fig.~4.2 and~4.3 in \cite{kok2017using}). Left: A snapshot of the platform for collecting real dataset  Right: Measurements from an accelerometer ($y_{\text{a},t}$, top), a gyroscope ($y_{\omega,t}$, middle) and a magnetometer ($y_{\text{m},t}$, bottom) for $100$ seconds of data collected with the IMU showed in the left figure.). 
  }
  \label{fig:oriEst-expSetup}
\end{figure}

We apply our algorithm for real dataset from \cite{kok2017using} (see Fig.~\ref{fig:oriEst-expSetup}). The data is collected from the Trivisio Colibri Wireless IMU \cite{Trivisio} with a logging rate of 100Hz. The reference measurement of the orientation is provided as ground truth from a motion capture equipment \cite{vicon} by tracking the optimal markers fixed to the sensor platform. The optical and IMU data has been time-synchronised and aligned beforehand.

\begin{figure}
    \centering
\includegraphics[scale=0.26]{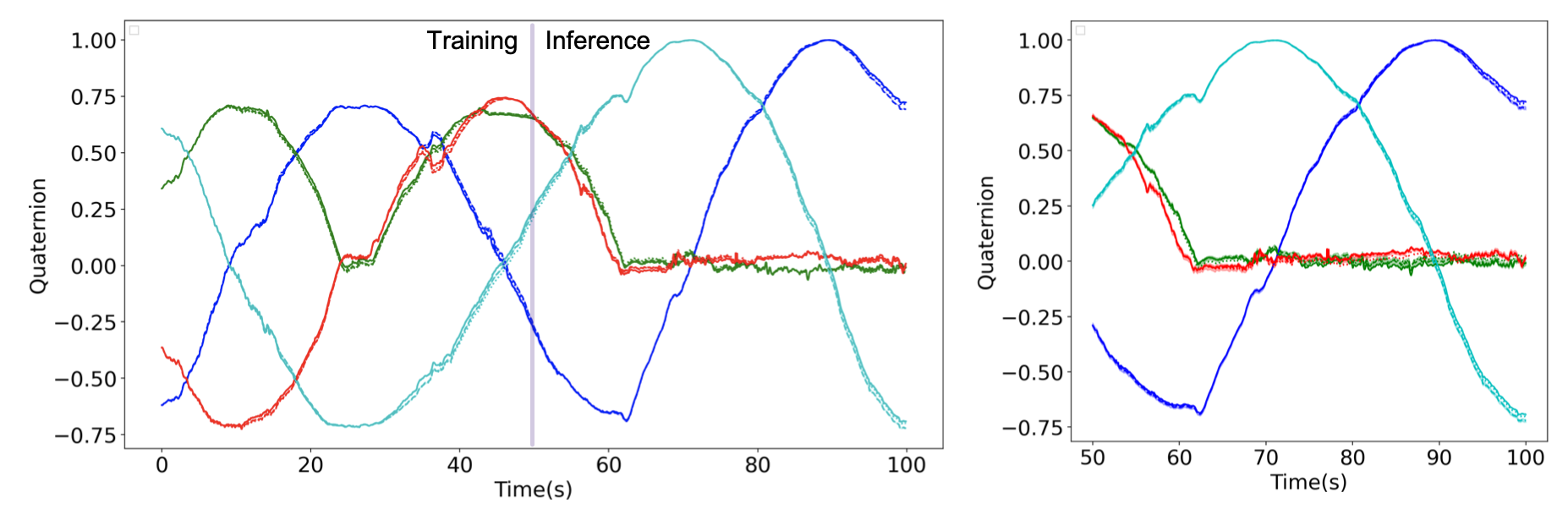}
    \vspace{-0.2cm}
    \caption{Quaternion for real data. The solid, dotted and dashed lines correspond to ground truth, integrated and estimated quaternion respectively. The left and right figures are corresponding to scenarios 1 and 2 respectively. In the right figure, the dashed line is obtained by averaging over $50$ independent trails.}
    \vspace{-0.2cm}
    \label{fig:real_data}
\end{figure}

The dataset is $100$ seconds long and split into training and inference dataset separately. The first half of the collected data is used for training and the rest for inference. In the training dataset, we randomly selected a consecutive sequence of a length of $1000$ samples as a training episode. We test in two scenarios (see Fig.~\ref{fig:oriEst-expSetup}): (1) inference including the training dataset from $t=0$ where the initial estimation $\hat{q}^\text{nb}_{t=0s}$ is simply the measurement; (2) inference without training dataset from $t=50s$ where $\hat{q}^\text{nb}_{t=50s}$ is normally distributed around the true initial orientation with a standard deviation of $0.01$. $50$ independent trials are performed. 
The estimation results are showed in Fig.~\ref{fig:real_data}. 

\begin{table}[h!]
\caption{RMSE of the orientation estimates}
\vspace{-0.2cm}
\label{tab:mean rmse}
\begin{center}
\small
\begin{tabular}{lcccc}
    \toprule
    {RMSE} & Yaw[$^{\circ}$] & Pitch[$^{\circ}$] & Roll[$^{\circ}$]\\
    \midrule
    Proposed algorithm  & 1.9423& 2.1048& 0.8353\\
    Extended Kalman Filter& 2.0411& 1.5272& 1.2488 \\
    Unscented Kalman Filter &20.1370& 20.3494& 38.6775 \\
    Complementary Filter & 1.2015& 1.3892& 0.8972 \\
    \bottomrule
    \vspace{-0.4cm}
\end{tabular}
\normalsize
\end{center}
\end{table}

In the second scenario, the second half of the dataset is also tested on the EKF, UKF and complementary filter methods. The estimation results are quantified and summarised in Table \ref{tab:mean rmse}. Since the initial estimations and the gyroscope measurements are relatively accurate, results indicate that our proposed algorithm has achieved comparable performance with the other three state-of-art algorithms.

\section{Conclusion}\label{sec: conclusion}
Orientation estimation using inertial sensors combined with a magnetometer is well studied, and many algorithms have been proposed. However, there hardly exist any algorithms with theoretical guarantees of estimation convergence. This paper proposes a reinforcement learning-based orientation estimation method and proves that its estimate error converges exponentially to a positive bound. The proposed method shows superior estimation performance compared with some well-known ones in terms of arbitrary estimation initialisation and adaptation to a drastic angular velocity profile.

\section{Acknowledgment}
We thank Rick Staa (TU Delft) for implementing the Algorithm~\ref{alg1} in TensorFlow \cite{abadi2016tensorflow}. We are grateful for the help and equipment provided by the UAS Technologies Lab, Artificial Intelligence and Integrated Computer Systems Division at the Department of Computer and Information Science, Link{\ "o}ping University, Sweden. We thank Gustaf Hendeby, Niklas Wahlstr{\ "o}m, Hanna Nyqvist and Manon Kok who collected the real data and allow us to use. This work is supported by Huawei, AnKobot and China Scholarship Council (No.202006890020).

\bibliographystyle{IEEEtran}

\end{document}